\documentclass[runningheads]{llncs}
\usepackage[T1]{fontenc}

\usepackage{amsmath}
\usepackage{url}
\usepackage[caption=false]{subfig}

\usepackage[table]{xcolor}
\usepackage{graphicx,verbatim}
\usepackage{booktabs}
\usepackage{multirow}
\begin{document}
\title{Ontology-Based Concept Distillation for Radiology Report Retrieval and Labeling}

\titlerunning{Ontology-Based Concept Distillation}

\author{Felix Nützel\inst{1}
\and
Mischa Dombrowski\inst{1}
\and
Bernhard Kainz\inst{1,2}
}
\authorrunning{F. Nützel et al.}
%
\institute{Department of Artificial Intelligence in Biomedical Engineering, Friedrich-Alexander-Universität Erlangen-Nürnberg, 91052 Erlangen, Germany
\email{\{felix.nuetzel,mischa.dombrowski,bernhard.kainz\}@fau.de}
\and
Department of Computing, Imperial College London, London SW7 2AZ, UK \\
}

\maketitle              
\begin{abstract}
Retrieval-augmented learning based on radiology reports has emerged as a promising direction to improve performance on long-tail medical imaging tasks, such as rare disease detection in chest X-rays.
Most existing methods rely on comparing high-dimensional text embeddings from models like CLIP or CXR-BERT, which are often difficult to interpret, computationally expensive, and not well-aligned with the structured nature of medical knowledge. 
We propose a novel, ontology-driven alternative for comparing radiology report texts based on clinically grounded concepts from the Unified Medical Language System (UMLS). Our method extracts standardised medical entities from free-text reports using an enhanced pipeline built on RadGraph-XL and SapBERT. These entities are linked to UMLS concepts (CUIs), enabling a transparent, interpretable set-based representation of each report. We then define a task-adaptive similarity measure based on a modified and weighted version of the Tversky Index that accounts for synonymy, negation, and hierarchical relationships between medical entities. This allows efficient and semantically meaningful similarity comparisons between reports. 
We demonstrate that our approach outperforms state-of-the-art embedding-based retrieval methods in a radiograph classification task on MIMIC-CXR, particularly in long-tail settings. Additionally, we use our pipeline to generate ontology-backed disease labels for MIMIC-CXR, offering a valuable new resource for downstream learning tasks. Our work provides more explainable, reliable, and task-specific retrieval strategies in clinical AI systems, especially when interpretability and domain knowledge integration are essential.
Our code is available at \url{https://github.com/Felix-012/ontology-concept-distillation}


\end{abstract}
\section{Introduction}

\noindent\textbf{Text Retrieval in Long-Tail Image Tasks}
Retrieving the most similar text captions or impressions has increasingly been used to design state-of-the art architectures in long-tail image data tasks, in particular in the biomedical domain.
Examples include clustering text captions to produce pseudo-classes with corresponding frequencies~\cite{dong_once_2024}, guiding the training of the classifier~\cite{zheng_large-scale_2024}, or to augment existing datasets for downstream tasks~\cite{dai_improving_2025}.

The dominant way to compare image captions is computing the cosine similarity between high-dimensional embeddings produced by multimodal models such as CLIP~\cite{clip}.
While such embeddings produce impressive results, they are difficult to interpret and they have not been designed for medical images, which poses problems in potentially high-stake environments, as often present in the biomedical domain.

\noindent\textbf{Ontology-based Text Similarity Measures}
An alternative approach is offered by ontology-based methods, which codify text appearances to unique entities and map the relations between those entities.
This way, a radiology report can be represented as a set of entities and the similarity between reports can be represented in terms of their set differences and relations, offering more insight into their relationship and providing a clear structure.
In the biomedical domain, the largest ontology is UMLS~\cite{UMLS2024AA}, which represents medical concepts in terms of identifiers called Concept Unique Identifiers (CUIs) and offers a myriad of additional information such as definitions, semantic types or relation types between CUIs.

Existing surveys of the field classify ontology-based approaches for measuring entity similarity into two main categories: path-based methods and Information Content (IC) methods~\cite{alonso_evaluation_2016,han_survey_2021}.
Some also include feature-based or set-based methods, which work via set comparisons, potentially augmented by information provided by the ontology~\cite{kulmanov_semantic_2021,lastra-diaz_reproducible_2019,pesquita_semantic_2009}.

IC-based methods derive the semantic similarity between CUIs based on their first shared ancestor (least common subsumer).
However, these methods do not scale well to sets, since they require to compute the least common subsumer for every pair between sets~\cite{kulmanov_semantic_2021}, resulting in $|set_1| \times |set_2|$ graph traversals per set, making them unfeasible for our downstream tasks.
A more efficient approach is to use the path distance between sets of entities. Applying these methods is not trivial, since the UMLS relation graph consists of thousands of connected components, making most nodes unreachable.
Another problem is that hierarchical granularities between disease classes are highly inconsistent, meaning that node-based distances rarely have a consistent semantic meaning.
The most efficient approach to compare entire sets of entities is to directly compare the set overlaps with metrics such as the Jaccard Index.
However, this family of approaches ignores the semantic relationships between entities and are therefore often too simplistic.

\noindent\textbf{Our Contribution}
In this work, we design a powerful entity extraction pipeline based on SapBERT and RadGraph-XL to construct a similarity measure for radiological reports that beats state-of-the-art text embeddings.
We achieve this by carefully incorporating semantic information from the UMLS knowledge base, such as synonyms or contradiction weighting.
Restricting comparisons to prototypical sets lets us exploit asymmetric similarity while keeping the measure symmetric. 
Assertions from RadGraph-XL, at both full-report and per-sentence levels, and multi-scale SapBERT contexts counter their extraction and assertion limits, yielding semantically rich CUI sets.
Combined with UMLS hierarchies, these sets produce ontology-backed labels for MIMIC-CXR~\cite{Johnson_2024_MIMIC_CXR_JPG}, boosting downstream performance on a curated test set for some disease classes.   

This way, researchers not only gain an insight into the relationship between label accuracy and image quality in long-tail retrieval scenarios, but also a dataset with the potential to substantially boost training results in the future.

In summary, our contributions include the following.
\begin{itemize}
    \item Introducing a powerful similarity metric for radiological reports based on UMLS entities and its corresponding entity retrieval pipeline
    \item Using our pipeline to generate ontology-backed disease labels for the MIMIC-CXR dataset
    \item Exploring the effect of asymmetric hierarchies in set similarity
    \item Demonstrating the effectiveness of our labels and similarity metric by training disease classifiers on long-tailed image datasets
\end{itemize}


\section{Method}
\begin{figure*}[htbp]
  \centering

  \subfloat[%
    \textbf{Left:} Our entity-extraction pipeline yields a CUI set per report. 
    \textbf{Middle:} A UMLS relation graph proposes candidate sets that we score with our distance function. 
    \textbf{Right:} Three UMLS tables (relations, semantic types, strings) feed the pipeline; coloured boxes mark their use.%
    \label{fig:pipeline}]%
  {%
    \begin{minipage}{\linewidth}
      \centering
      \includegraphics[width=.85\linewidth]{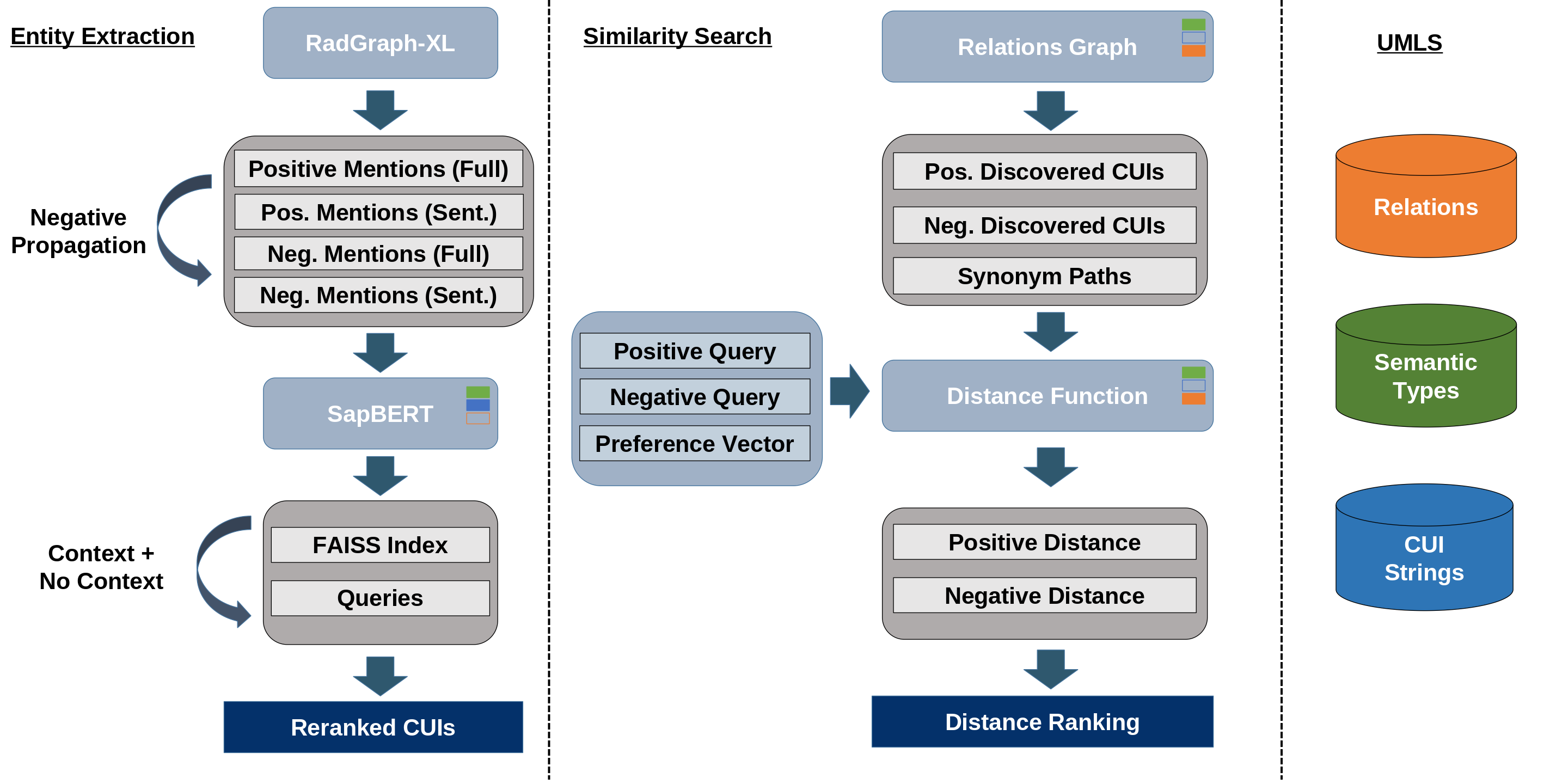}
    \end{minipage}
  } \par 

  \subfloat[%
    We reconstruct the original dataset with ten similarity-search queries per disease. 
    The retrieved set trains an image classifier that we compare against one trained on the balanced reference set.%
    \label{fig:retrieval}]%
  {%
    \begin{minipage}{\linewidth}
      \centering
      \includegraphics[width=.8\linewidth]{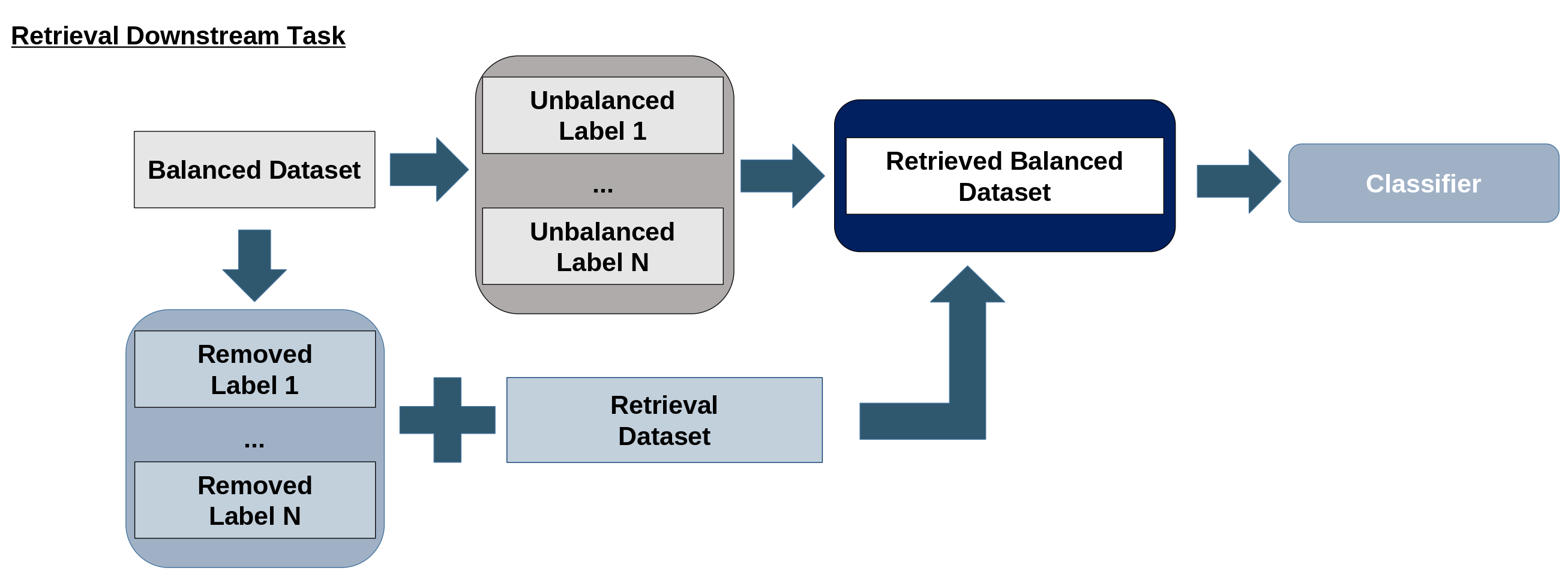}
    \end{minipage}
  } \par

  \subfloat[%
    \textbf{First stage:} Report CUIs map to the label set; unused CUIs fall back to their parent. 
    \textbf{Second stage:} Baseline and new labels become CUI sets, and the closest label to each report’s set is chosen.%
    \label{fig:labeler}]%
  {%
    \begin{minipage}{\linewidth}
      \centering
      \includegraphics[width=.85\linewidth]{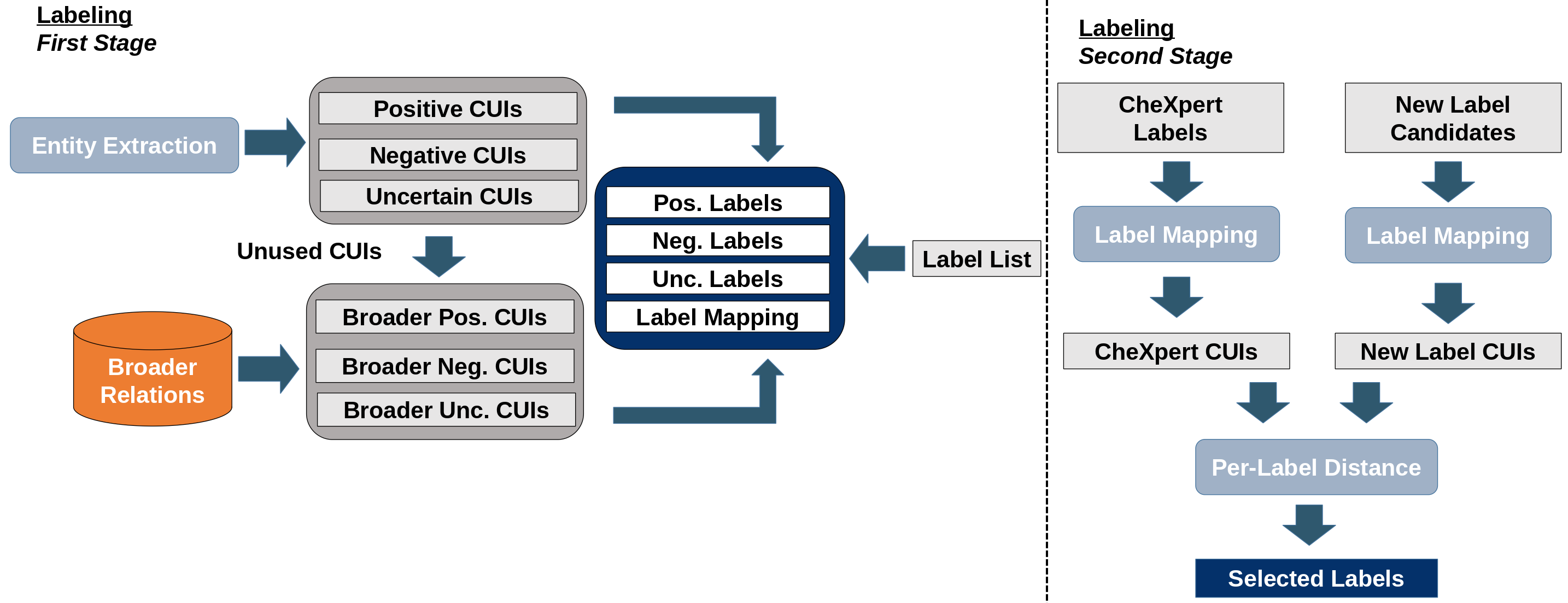}
    \end{minipage}
  } 

  \caption{Our method comprises four modules: entity extraction, similarity search, the retrieval experiment, and the labelling procedure.}
  \label{fig:method_overview}
\end{figure*}
As illustrated in Figure~\ref{fig:method_overview}, our proposed approach comprises four main components: the entity extraction pipeline, the similarity search framework, the downstream retrieval task, and the labeling procedure.

\noindent\textbf{Entity Extraction}
First, we extract mentions of medical entities from MIMIC-CXR radiology reports via RadGraph-XL~\cite{delbrouck-etal-2024-radgraph}. 
RadGraph-XL can detect the presence, absence, or uncertainty of the entity type \emph{Anatomy} (e.g., "Lung") and the entity type \emph{Observation} (e.g., "Pneumonia"). 
Typically, anatomies are different from observations in that they do not inherit negations from related observations, since negating an anatomic region of the radiograph is usually not meaningful. 
However, this leads to impressions such as "No consolidations in the lung" to simply consist of the anatomical entity "lung". For most downstream tasks, such as anomaly detection, this is undesirable.
Therefore, we propagate negations to all related entities, instead of only considering observations, and negate anatomies with no corresponding positive observation.
We provide both the entire report, as well as a list of its sentences to RadGraph-XL, since having both contexts leads to more accurate entity and assertion extraction in our experiments.
However, RadGraph-XL produces inaccurate assertions for short noun phrases with a single modifier, since it is biased towards negating such samples.
Therefore, we apply a straightforward rule-based assertion in these simple edge cases.

To link the entities to CUIs, we built an index for similarity search by embedding each CUI string via SapBERT~\cite{liu2021learning} fine-tuned on the UMLS dataset.
We limit the set of vocabularies to SNOMEDCT-US, a comprehensive medical terminology used to standardize working with electronic health data, since mixing vocabularies results to conflicting CUIs, or ones that share no relations in the UMLS ontology.
Then, we use SapBERT to embed each isolated entity and each entity context.
This way, we capture both more general concepts described by the isolated mentions and the more specific or context-aware concepts captured by considering all mentions that modify each other.
To find the best candidate for retrieval, we perform a top-128 cosine similarity search on the precomputed index.
We filter the candidates based on related semantic types.
We limited observations to four semantic types, and anatomies to five semantic types.
Among the remaining CUIs, we link the one with the highest similarity score to the respective entity.
This way, each radiology report can be represented as a set of CUIs.

\noindent\textbf{Knowledge Graph Construction}
We construct a cyclic, undirected graph, based on the UMLS relations concerned with hierarchy and semantic closeness.
This graph serves two purposes: we use it to find potential report candidates in the dataset by performing a Breadth-First Search of depth $n (=10)$  and retrieving all report sets that overlap with the discovered graph subset.
The discovery process has no larger impact on the results of our method and can be left out.
However, it can be useful for performance gains for very large datasets if $|discovered| \ll |dataset|$.
The second use case is synonym discovery. If a CUI from the target set can reach a CUI in a candidate set by only using synonym relations, this CUI is considered a synonym and is added to the target set when comparing these two sets.
We do not directly use the semantic distances present in the graph, since this caused a computational overhead with no performance gains in our experiments.

\noindent\textbf{Preference-based CUI Set Distance}
To evaluate the distance between two CUI sets, we propose a modified version of the Tversky Index~(Eq.~\ref{eq:tversky}), coupled with a preference vector as parameter.

\begin{equation}
\label{eq:tversky}
    D_{\alpha,\beta}(A,B)\;=\;
\frac{\lvert A\cap B\rvert}
     {\lvert A\cap B\rvert
      \;+\;\alpha\,\lvert A\setminus B\rvert
      \;+\;\beta\,\lvert B\setminus A\rvert}
\end{equation}
The Tversky Index is a generalization of set similarity metrics such as the Jaccard Index or the Dice-Sørensen coefficient. 
In particular, if the parameters $\alpha = 1 $ and $\beta = 1$, then the Tversky Index is equal to the Jaccard Index, and if $\alpha = 0.5 $ and $\beta = 0.5$, then it is equal to the Dice-Sørensen coefficient.
Tversky found that prototypical concepts are usually less similar to specific concepts than vice-versa~\cite{Tversky1977Features}. This means that despite the fact that most similarity metrics are symmetrical, the human notion of similarity is actually asymmetric. This concept can be easily extended to retrieval tasks, e.g., retrieving the term "whale" based on the term "animal" would be less expected than the other way around.
However, many tasks require a symmetrical measure, which is why we use a symmetric reformulation that dynamically decides the prototype by inspecting which set difference is larger~\cite{jimenez-etal-2013-softcardinality}:
\begin{equation}
D^{sym}_{\alpha,\beta}(A,B)
=\frac{|A\cap B|}
       {|A\cap B|
        +\beta\!\left(
           \alpha\,a + (1-\alpha)\,b
        \right)},
\qquad
\begin{aligned}
a &= \min\!\bigl(|A\setminus B|,\;|B\setminus A|\bigr),\\[4pt]
b &= \max\!\bigl(|A\setminus B|,\;|B\setminus A|\bigr).
\end{aligned}
\end{equation}

To give the prototype the deciding weight and to simplify the equation, we choose $\alpha = 0$, resulting in the simplified version:

\begin{equation}
D^{prot}_\beta(A,B)
=\frac{|A\cap B|}
       {|A\cap B|+\beta\,b},
\qquad
b=\max\!\bigl(|A\setminus B|,\;|B\setminus A|\bigr).
\end{equation}

whereas $\beta = 1$. In our experiments we use this equation, with the Jaccard Index $(\alpha=0.5, \beta=1)$ as an additional baseline.

If a target set negates an observation present in a candidate set, we call this a contradiction.
Intuitively, contradicting sets should be less similar than a non-contradicting counterpart, which is why we added an additional contradiction term to the Tversky Index. Therefore, we heuristically add a contradiction term $C$, the number of contradictions to our equation.

Semantic similarity is not task-agnostic, despite most similarity measures being defined this way. 
As an example, consider the impressions "lower left pneumothorax", "lower left pneumonia", and "upper right pneumothorax".
If the task is disease localization for example, the first and second impression would result in more similar bounding boxes, so they would be the closest pair in this case. 
However, if the task is disease detection, the first and third impression would be the closest, since they describe the same disease. 
No ordering is strictly more correct than the other without considering the task at hand.
Additionally, as the above example shows, the type of information that might be relevant for semantic similarity is mostly easy to decide.
One only needs to consider the relevance of the four observation types and the five anatomical types.
Therefore, we propose adding a preference vector that filters and/or ranks the required semantic types based on the task.
In the most straightforward manner, it can be used to filter out semantic types by setting them to 0.
In addition, one can enforce an order to the semantic types, by assigning values in descending order to them.
The filter is applied to each component of our measure, so the relative distances are only influenced by the proportionality of the values.
This results in our final distance equation:
\begin{equation}
    D = \frac{\sum{W_{A\cap B}}}{\sum{W_{A \cap B}} + \max(\sum{W_{A\setminus B}}, \sum{W_{B\setminus A}}) + \sum{W_{contr}}}, 
\end{equation}
with $W_{set}$ being the preference weights of the corresponding set and $W_{contr}$ being the weights of the contradicting CUIs.

\noindent\textbf{Downstream Task}
To evaluate the performance of our retrieval method, we use a surrogate DenseNet-121 classifier following \cite{rajpurkar_chexnet_2017} to evaluate the performance of our method, as can be seen in Figure~\ref{fig:retrieval}. 
Test results are reported on the official test split, which comprises 887 samples verified by radiologists. 
We restrict our analysis to the eight most frequent disease classes, as the remaining classes occur too infrequently to allow for reliable evaluation.
The baseline method uses a balanced dataset with 706 randomly selected samples.

To mimic a long-tail retrieval task we take ten entries per class from the balanced train dataset $B$, resulting in our query dataset $Q$. The removed diseases then are combined with a holdback dataset to form our retrieval dataset $R$.
This results in the relationship $(B \setminus Q )\subset R$.
In round-robin manner, we perform similarity search for one of the queries of each disease at a time. 
This allows us to simulate one long-tail retrieval scenario for each disease using a single run.

\noindent\textbf{Labeling Setup}
To improve on the automatically generated labels provided by the MIMIC-CXR dataset, we use the CUI set representations to provide more accuracte labels which in turn results in better performance when using this dataset for downstream tasks. 
Our approach is as visualized in Figure~\ref{fig:labeler}, consisting of two phases: We first use our entity extraction pipeline to associate each radiology report with a set of CUIs.
Then, we match the CUI name with each of the labels, resulting in a mapping between each label and a corresponding set of CUIs.
Given the surface CUI and the isolated CUI of a mention, only the label with the best score is considered, so a single mention can only cause one label.
If a CUI could not be matched to a label, its parent CUI is retrieved, which in turn is matched with the labels.
In accordance with CheXpert, the labels are marked with 1, 0 and -1 according to their Radgraph-XL assertions, resulting in our preliminary labels.

In the second phase, the CUI sets corresponding to the original labels and our preliminary labels are compared to the report CUIs.
Since we are primarily interested in matches regardless of the size of the label sets, we use the asymmetric containment index $|R \cap L|/|R|$ for a report set $R$ and a label set $L$ as our similarity measure here. 
Given the two label sets $L_{new}$ and $L_{old}$, we evaluate this measure for $L_{old}$, $L_{new}$, $L_{new} \cap L_{old}$ and $L_{new} \cup L_{old}$. 
Among these, the most similar set is chosen as the final label for the report. 

\begin{table}[htbp]
  \centering
  \footnotesize
  \caption{Area-under-ROC (AUROC) for each trained classifier averaged over six runs (mean±standard deviation) on the baseline dataset (706 images) and datasets retrieved by different retrieval methods (using prompts from 10 images).}
  \label{tab:auroc_classifier}
  \resizebox{\textwidth}{!}{%
  \begin{tabular}{lccccccccc}
    \toprule
    \textbf{Model} &
    \textbf{No~Finding} &
    \textbf{Atelectasis} &
    \textbf{Cardio\-megaly} &
    \textbf{Consoli\-dation} &
    \textbf{Edema} &
    \textbf{Pleural Eff.} &
    \textbf{Pneumonia} &
    \textbf{Pneumothorax} &
    \textbf{Mean} \\
    \midrule
    Balanced Dataset & 0.849±.003 & 0.772±.014 & 0.784±.011 & 0.776±.015 & 0.862±.007 & 0.826±.004 & 0.739±.009 & 0.831±.014 & \textbf{0.805±.006} \\
    \midrule
    \multicolumn{10}{l}{\textit{Embedding-based methods}}\\[0.25em]
    CXR-BERT~\cite{cxr-bert} Retrieval & 0.825±.006 & 0.713±.022 & 0.648±.016 & 0.590±.014 & 0.791±.014 & 0.775±.015 & 0.620±.023 & 0.726±.025 & \textbf{0.711±.006} \\
    CLIP~\cite{clip} Retrieval & 0.815±.009 & 0.691±.019 & 0.661±.029 & 0.570±.029 & 0.774±.019 & 0.750±.013 & 0.656±.023 & 0.696±.012 & \textbf{0.702±.010} \\
    Bio-ClinicalBERT~\cite{biobert} & 0.821±.012 & 0.668±.032 & 0.646±.034 & 0.611±.022 & 0.758±.032 & 0.694±.025 & 0.632±.029 & 0.668±.007 & \textbf{0.687±.010} \\
    \midrule
    \multicolumn{10}{l}{\textit{Ontology-based methods}}\\[0.25em]
    Path Retrieval &
      0.819 & 0.519 & 0.726 & 0.591 &
      0.645 & 0.510 & 0.555 & 0.537 &
      \textbf{0.613} \\
    \rowcolor{gray!15}Jaccard Retrieval (Ours) & 0.829±.002 & 0.748±.011 & 0.701±.019 & 0.654±.013 & 0.835±.013 & 0.788±.014 & 0.661±.021 & 0.696±.027 & \textbf{0.739±.007} \\
    \rowcolor{gray!15}Prototypical Retrieval (Ours) & 0.824±.009 & 0.754±.012 & 0.716±.014 & 0.657±.027 & 0.830±.014 & 0.785±.013 & 0.665±.021 & 0.701±.023 & \textbf{0.741±.008} \\
    \bottomrule
  \end{tabular}
  }
\end{table}

\section{Results \& Discussion}
\noindent\textbf{Retrieval Results}
In Table~\ref{tab:auroc_classifier}, we see the training results of the classifer model based on the dataset it received.
Among the retrieval methods, ours improves on CXR-BERT in five out of eight disease classes, on CLIP on seven out of eight classes and on the others on all disease classes, achieving the highest mean AUROC score among all of them.
This demonstrates the strength of our approach, and that text embeddings are not necessarily the best method for long-tail data retrieval tasks.
Using a simple path-based approach instead of a set difference approach yields the worst results among the tested base lines, affirming our approach compared to related set comparison options.

\noindent\textbf{Labeling Results}\begin{table}[htbp]
  \centering
  \footnotesize
  \caption{AUROC for each trained classifier averaged over six runs (mean±standard deviation) on the baseline CheXpert labels for MIMIC-CXR, the labels generated by our baselines and the labels generated by us.}
  \label{tab:labels}
  \resizebox{\textwidth}{!}{%
  \begin{tabular}{lccccccccc}
    \toprule
    \textbf{Model} &
    \textbf{No~Finding} &
    \textbf{Atelectasis} &
    \textbf{Cardio\-megaly} &
    \textbf{Consoli\-dation} &
    \textbf{Edema} &
    \textbf{Pleural Eff.} &
    \textbf{Pneumonia} &
    \textbf{Pneumothorax} &
    \textbf{Mean} \\
    \midrule
    \multicolumn{10}{l}{\textit{Baseline}}\\[0.25em]
    CheXpert~\cite{irvin2019chexpert} & 0.849±.003 & 0.772±.014 & 0.784±.011 & 0.776±.015 & 0.862±.007 & 0.826±.004 & 0.739±.009 & 0.831±.014 & \textbf{0.805±.006} \\
    \midrule
    \multicolumn{10}{l}{\textit{LLM-based methods}}\\[0.25em]
    LLama 3.3~\cite{grattafiori2024llama3herdmodels} & 0.834±.006 & 0.749±.020 & 0.773±.013 & 0.764±.023 & 0.854±.007 & 0.578±.026 & 0.681±.014 & 0.818±.012 & \textbf{0.756±.011} \\
    DeepSeek 2~\cite{deepseekv2} & 0.836±.005 & 0.743±.013 & 0.777±.012 & 0.770±.022 & 0.849±.011 & 0.552±.033 & 0.668±.019 & 0.819±.009 & \textbf{0.752±.010} \\
    MedGemma~\cite{medgemma-hf} & 0.831±.007 & 0.729±.015 & 0.759±.018 & 0.758±.015 & 0.833±.008 & 0.567±.035 & 0.635±.012 & 0.801±.014 & \textbf{0.739±.007} \\
    \midrule
    \multicolumn{10}{l}{\textit{Ours}}\\[0.25em]
    Without Retrieval & 0.833±.008 & 0.746±.015 & 0.769±.011 & 0.752±.016 & 0.832±.006 & 0.823±.017 & 0.677±.025 & 0.820±.008 & \textbf{0.782±.010} \\
      Jaccard Retrieval & 0.843±.005 & 0.755±.003 & 0.777±.016 & 0.769±.014 & 0.845±.008 & 0.827±.017 & 0.737±.011 & 0.82±.017 & \textbf{0.797±.009} \\
    \rowcolor{gray!15} Containment Retrieval & 0.840±.013 & 0.772±.006 & 0.782±.016 & 0.772±.023 & 0.856±.010 & 0.826±.008 & 0.734±.020 & 0.832±.020 & \textbf{0.802±.012} \\
    \bottomrule
  \end{tabular}
  }
\end{table}

When inspecting Table~\ref{tab:labels}, it is evident that the baseline CheXpert labels still perform considerably better than modern language models, despite their use of advanced techniques such as chain-of-thought by DeepSeek for example.
Many reports in MIMIC-CXR, such as reports only comparing to unknown references, cannot be conclusively labeled correctly, so a perfect labeling is not possible using text-based approaches.
However, there are still some glaring mislabelings used by CheXpert, which our retrieval method could remedy for some disease classes, resulting in better classification performance for Pleural Effusion and Pneumothorax.
As was the case in our retrieval downstream task, our ablation result for using the Jaccard Index highlights the benefit when using an asymmetrical measure in the labeling context as well.

\noindent\textbf{Limitations}
While we provided baselines for the most important state-of-the-art similarity search approaches, there are still many methods we could not include in our scope, which is in particular true for the large corpus of ontology-based retrieval algorithms.
Adding an IC metric baseline was not viable, due to performance issues.
However, since embedding-based models are used for the vast majority of retrieval tasks, we believe that our selection still forms a meaningful basis of comparison for our method.

\section{Conclusion}
In this work, we have shown how to use current models to create a competetive entity extraction pipeline and how to effectively evaluate the extracted entities using similarity measures.
Using a long-tail image classification task, we attained superior results compared to popular embedding models.
Our results highlight the significant potential of ontology-based approaches in multimodal long-tail retrieval tasks, which have shown to be an important building blocks in recent works, in particular in the biomedical domain. 
Future research should aim to apply our retrieval method and generated MIMIC-CXR labels in more advanced down stream tasks.
Additionally, while improvements have been made in this field, entity extraction and linking models still need further improvement, as state-of-the-art models such as RadGraph-XL still fail even the most basic cases.

\newpage
\bibliographystyle{splncs04}
\bibliography{main}

\begin{thebibliography}{10}
\providecommand{\url}[1]{\texttt{#1}}
\providecommand{\urlprefix}{URL }
\providecommand{\doi}[1]{https://doi.org/#1}

\bibitem{alonso_evaluation_2016}
Alonso, I., Contreras, D.: Evaluation of semantic similarity metrics applied to the automatic retrieval of medical documents: {An} {UMLS} approach. Expert Systems with Applications  \textbf{44},  386--399 (Feb 2016). \doi{10.1016/j.eswa.2015.09.028}, \url{https://www.sciencedirect.com/science/article/pii/S0957417415006569}

\bibitem{biobert}
Alsentzer, E., et~al.: Publicly available clinical {BERT} embeddings. In: Proceedings of the 2nd Clinical Natural Language Processing Workshop. pp. 72--78. Association for Computational Linguistics, Minneapolis, Minnesota, USA (Jun 2019). \doi{10.18653/v1/W19-1909}, \url{https://www.aclweb.org/anthology/W19-1909}

\bibitem{cxr-bert}
Boecking, B., et~al.: Making the most of text semantics to improve biomedical vision--language processing. In: Avidan, S., Brostow, G., Ciss{\'e}, M., Farinella, G.M., Hassner, T. (eds.) Computer Vision -- ECCV 2022. pp. 1--21. Springer Nature Switzerland, Cham (2022)

\bibitem{dai_improving_2025}
Dai, F., et~al.: Improving {AI} models for rare thyroid cancer subtype by text guided diffusion models. Nature Communications  \textbf{16}(1), ~4449 (May 2025). \doi{10.1038/s41467-025-59478-8}, \url{https://www.nature.com/articles/s41467-025-59478-8}, publisher: Nature Publishing Group

\bibitem{deepseekv2}
DeepSeek-AI: Deepseek-v2: A strong, economical, and efficient mixture-of-experts language model (2024)

\bibitem{delbrouck-etal-2024-radgraph}
Delbrouck, J.B., et~al.: {R}ad{G}raph-{XL}: A large-scale expert-annotated dataset for entity and relation extraction from radiology reports. In: Ku, L.W., Martins, A., Srikumar, V. (eds.) Findings of the Association for Computational Linguistics: ACL 2024. pp. 12902--12915. Association for Computational Linguistics, Bangkok, Thailand (Aug 2024). \doi{10.18653/v1/2024.findings-acl.765}, \url{https://aclanthology.org/2024.findings-acl.765/}

\bibitem{dong_once_2024}
Dong, F., et~al.: Once read is enough: Domain-specific pretraining-free language models with cluster-guided sparse experts for long-tail domain knowledge. In: Globerson, A., Mackey, L., Belgrave, D., Fan, A., Paquet, U., Tomczak, J., Zhang, C. (eds.) Advances in Neural Information Processing Systems. vol.~37, pp. 88956--88980. Curran Associates, Inc. (2024)

\bibitem{medgemma-hf}
Google: Medgemma hugging face. \url{https://huggingface.co/collections/google/medgemma-release-680aade845f90bec6a3f60c4} (2025), accessed: [Insert Date Accessed, e.g., 2025-05-20]

\bibitem{grattafiori2024llama3herdmodels}
Grattafiori, A., et~al.: The llama 3 herd of models (2024), \url{https://arxiv.org/abs/2407.21783}

\bibitem{han_survey_2021}
Han, M., Zhang, X., Yuan, X., Jiang, J., Yun, W., Gao, C.: A survey on the techniques, applications, and performance of short text semantic similarity. Concurrency and Computation: Practice and Experience  \textbf{33}(5),  e5971 (2021). \doi{10.1002/cpe.5971}, \url{https://onlinelibrary.wiley.com/doi/abs/10.1002/cpe.5971}, \_eprint: https://onlinelibrary.wiley.com/doi/pdf/10.1002/cpe.5971

\bibitem{irvin2019chexpert}
Irvin, J., et~al.: Chexpert: A large chest radiograph dataset with uncertainty labels and expert comparison. In: Thirty-Third AAAI Conference on Artificial Intelligence (2019)

\bibitem{jimenez-etal-2013-softcardinality}
Jimenez, S., Becerra, C., Gelbukh, A.: {SOFTCARDINALITY}-{CORE}: Improving text overlap with distributional measures for semantic textual similarity. In: Diab, M., Baldwin, T., Baroni, M. (eds.) Second Joint Conference on Lexical and Computational Semantics (*{SEM}), Volume 1: Proceedings of the Main Conference and the Shared Task: Semantic Textual Similarity. pp. 194--201. Association for Computational Linguistics, Atlanta, Georgia, USA (Jun 2013), \url{https://aclanthology.org/S13-1028/}

\bibitem{Johnson_2024_MIMIC_CXR_JPG}
Johnson, A., Lungren, M., Peng, Y., Lu, Z., Mark, R., Berkowitz, S., Horng, S.: Mimic-cxr-jpg - chest radiographs with structured labels (2024). \doi{https://doi.org/10.13026/jsn5-t979}

\bibitem{kulmanov_semantic_2021}
Kulmanov, M., Smaili, F.Z., Gao, X., Hoehndorf, R.: Semantic similarity and machine learning with ontologies. Briefings in Bioinformatics  \textbf{22}(4),  bbaa199 (Jul 2021). \doi{10.1093/bib/bbaa199}, \url{https://doi.org/10.1093/bib/bbaa199}

\bibitem{lastra-diaz_reproducible_2019}
Lastra-Díaz, J.J., Goikoetxea, J., Hadj~Taieb, M.A., García-Serrano, A., Ben~Aouicha, M., Agirre, E.: A reproducible survey on word embeddings and ontology-based methods for word similarity: {Linear} combinations outperform the state of the art. Engineering Applications of Artificial Intelligence  \textbf{85},  645--665 (Oct 2019). \doi{10.1016/j.engappai.2019.07.010}, \url{https://www.sciencedirect.com/science/article/pii/S0952197619301745}

\bibitem{liu2021learning}
Liu, F., Vuli{\'c}, I., Korhonen, A., Collier, N.: Learning domain-specialised representations for cross-lingual biomedical entity linking. In: Proceedings of ACL-IJCNLP 2021. pp. 565--574 (Aug 2021)

\bibitem{UMLS2024AA}
{National Library of Medicine (US)}: {UMLS Knowledge Sources [dataset on the Internet]}. Bethesda (MD): National Library of Medicine (US) (May 2024), \url{http://www.nlm.nih.gov/research/umls/licensedcontent/umlsknowledgesources.html}, [cited 2024-07-15]

\bibitem{pesquita_semantic_2009}
Pesquita, C., Faria, D., Falcão, A.O., Lord, P., Couto, F.M.: Semantic {Similarity} in {Biomedical} {Ontologies}. PLOS Computational Biology  \textbf{5}(7),  e1000443 (Jul 2009). \doi{10.1371/journal.pcbi.1000443}, \url{https://journals.plos.org/ploscompbiol/article?id=10.1371/journal.pcbi.1000443}, publisher: Public Library of Science

\bibitem{clip}
Radford, A., et~al.: Learning transferable visual models from natural language supervision. In: ICML (2021)

\bibitem{rajpurkar_chexnet_2017}
Rajpurkar, P., et~al.: {CheXNet}: {Radiologist}-{Level} {Pneumonia} {Detection} on {Chest} {X}-{Rays} with {Deep} {Learning} (2017)

\bibitem{Tversky1977Features}
Tversky, A.: Features of similarity. Psychological Review  \textbf{84}(4),  327--352 (1977). \doi{10.1037/0033-295X.84.4.327}

\bibitem{zheng_large-scale_2024}
Zheng, Q., et~al.: Large-scale long-tailed disease diagnosis on radiology images. Nature Communications  \textbf{15}(1),  10147 (Nov 2024). \doi{10.1038/s41467-024-54424-6}, \url{https://doi.org/10.1038/s41467-024-54424-6}

\end{thebibliography}
\end{document}